\documentclass[10pt]{article}
\usepackage{newtxtext}
\usepackage{newtxmath}
\usepackage{gensymb}

\DeclareFontEncoding{LS1}{}{}
\DeclareFontSubstitution{LS1}{stix}{m}{n}
\DeclareRobustCommand{\diameter}{%
  \text{\usefont{LS1}{stixscr}{m}{n}\symbol{"60}}%
}

\usepackage[T1]{fontenc}

\usepackage{graphicx}
\usepackage{url,hyperref}

\usepackage{multicol}
\usepackage{float}
\usepackage[caption=false]{subfig}

\usepackage[numbers]{natbib}
\usepackage{geometry}
\providecommand{\keywords}[1]
{
	\small
	\textbf{\textit{Keywords---}} #1
}

\usepackage{xcolor}
\hypersetup{
	colorlinks,
	linkcolor={red!50!black},
	citecolor={blue!50!black},
	urlcolor={blue!80!black}
}

\begin{document}
\title{Improved mirror ball projection for more accurate merging of multiple camera outputs and process monitoring}

\author{Wladislav Artsimovich\thanks{DMG MORI CO. LTD.,	2-3-23 Shiomi, Koto-ku, Tokyo 135-0052 Japan}\hspace{2mm}\thanks{Fraunhofer-Institut für Werkstoff- und Strahltechnik, Winterbergstr. 28, 01277 Dresden, Germany}\hspace{2mm}\thanks{Berufsakademie Sachsen, Staatliche Studienakademie Dresden, Hans-Grundig-Straße 25, 01307 Dresden, Germany}\hspace{5mm} Yoko Hirono\footnotemark[1]\\
	\href{mailto:wladislav.artsimovich@dmgmori.co.jp}{wladislav.artsimovich@dmgmori.co.jp}\hspace{6mm}\href{mailto:wladislav.artsimovich@dmgmori.co.jp}{yo-hirono@dmgmori.co.jp}\\
}

\maketitle

\begin{abstract}
	Using spherical mirrors in place of wide-angle cameras allows for cost-effective monitoring of manufacturing processes in hazardous environment, where a camera would normally not operate. This includes environments of high heat, vacuum and strong electromagnetic fields. Moreover, it allows the layering of multiple camera types (e.g., color image, near-infrared, long-wavelength infrared, ultraviolet) into a single wide-angle output, whilst accounting for the different camera placements and lenses used. Normally, the different camera positions introduce a parallax shift between the images, but with a spherical projection as produced by a spherical mirror, this parallax shift is reduced, depending on mirror size and distance to the monitoring target.

	This paper introduces a variation of the 'mirror ball projection', that accounts for distortion produced by a perspective camera at the pole of the projection. Finally, the efficacy of process monitoring via a mirror ball is evaluated.
\end{abstract}


\begin{figure}[h]
	\centering
	\subfloat[Color live stream]{
		\includegraphics[width=0.23\textwidth]{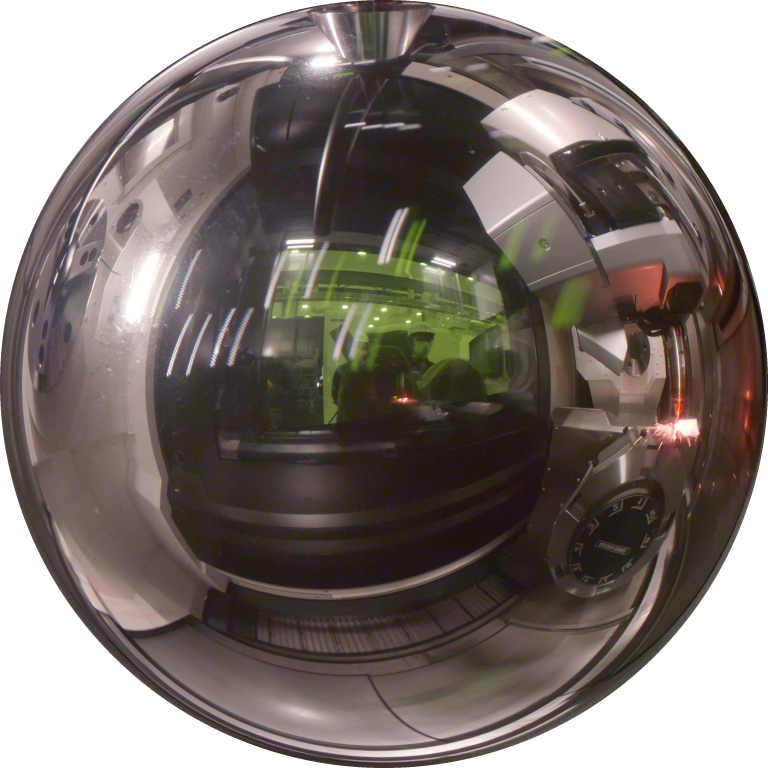}
		\label{fig:color}
	}
	\hfill
	\subfloat[Infrared live stream]{
		\includegraphics[width=0.23\textwidth]{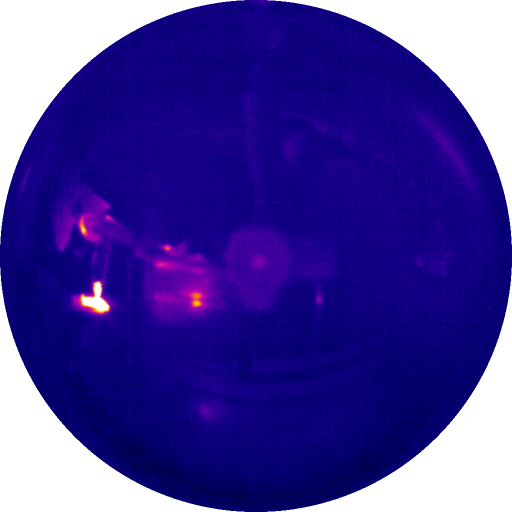}
		\label{fig:thermal}
	}
	\hfill
	\subfloat[Projected color view]{
		\includegraphics[width=0.23\textwidth]{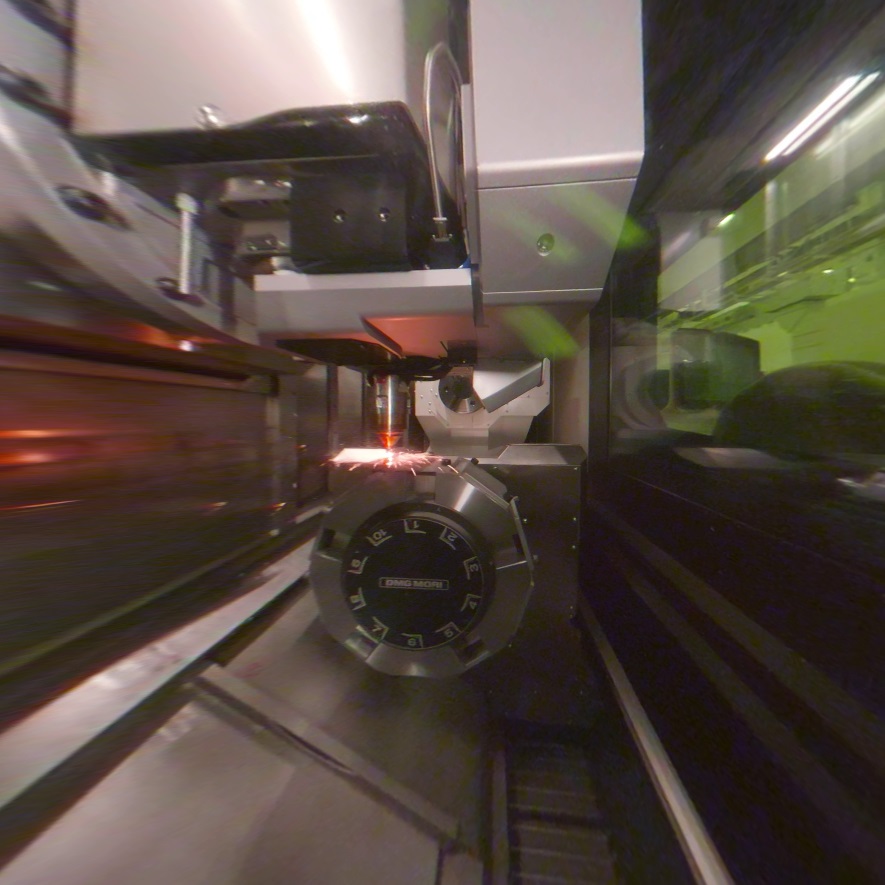}
		\label{fig:color_proj}
	}
	\hfill
	\subfloat[Projected thermal view]{
		\includegraphics[width=0.23\textwidth]{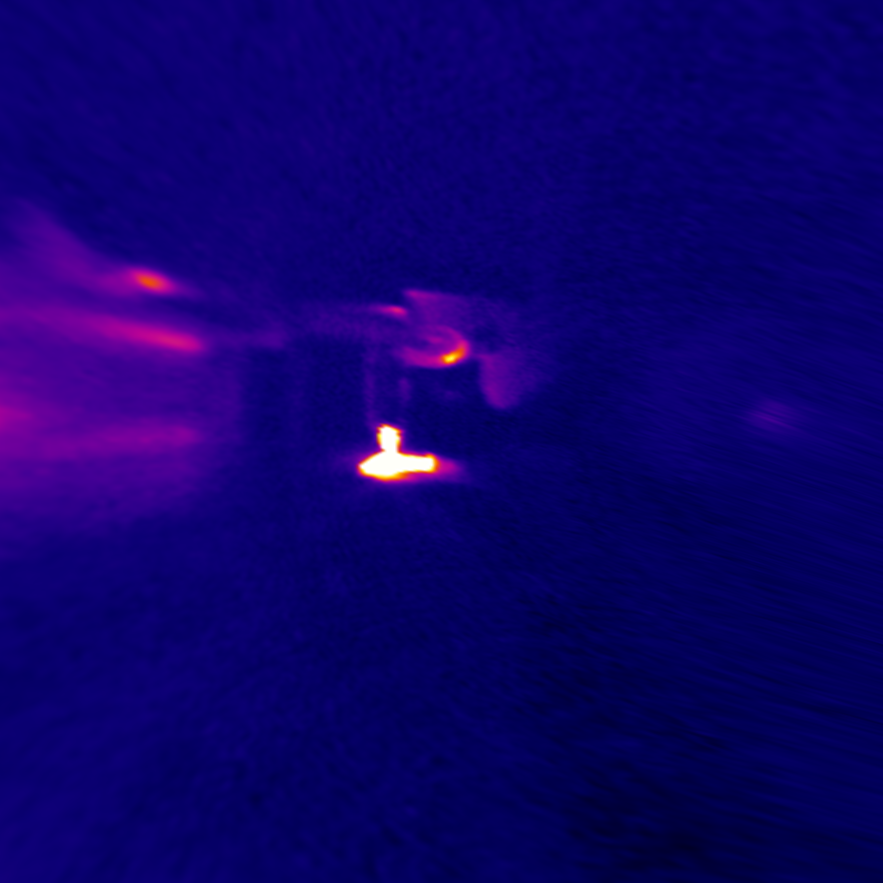}
		\label{fig:thermal_proj}
	}

	\caption{Multi-source mirror ball live stream of a laser cladding process}
	\label{fig:mirror-process}
\end{figure}

\begin{multicols}{2}
	\section{Introduction}

	Spherical mirror projections are used in a wide variety of applications as a substitute for wide-angle or panoramic camera systems. This extends to the field of robotics with 'omnidirectional sensors' as used in \cite{omnidirectional} and to the field of computer graphics with various 'image-based lighting' techniques (\cite{lightprobe}) or novel ways of displaying and interacting with panoramic content (\cite{dome}). A less explored use case is process monitoring.

	The mirror ball can be made to withstand extreme conditions at relatively low cost. For instance, stainless steel ball bearings are a mass-produced commodity, that can survive high heat environments, whilst producing a mirror ball projection. Moreover, multiple cameras can be pointed at the same mirror sphere, all identically encoding the full 360° environment, regardless of camera position, at least mathematically. This allows for merging of multiple camera outputs into a single video feed.

	A mirror ball is mathematically the simplest spherical mirror and projects the full 360° environment onto a 2D plane, when captured with an orthographic camera. Whilst an orthographic camera can be achieved in the real-world with the use of a telecentric lens, almost all cameras are perspective ones. Distortion at the pole-point of the projection is introduced by merely approximating an orthographic camera when using a perspective one.

	\begin{figure*}
		\centering
		\subfloat[Scene setup]{
			\includegraphics[height=0.21\textwidth]{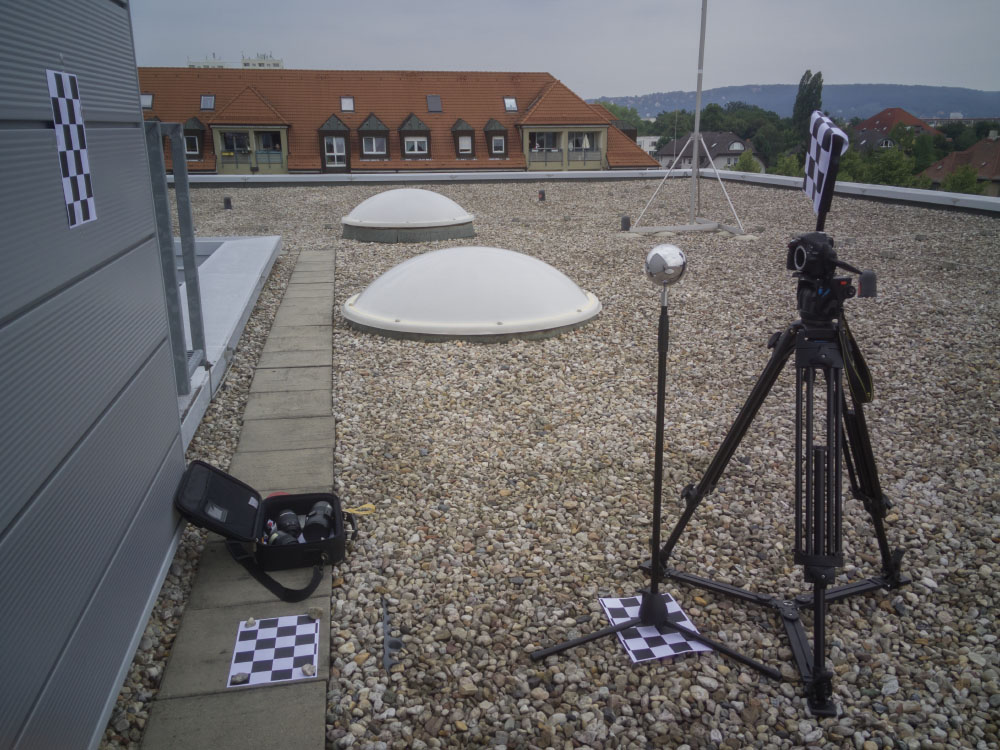}
			\label{fig:ball_setup}
		}
		\hfill
		\subfloat[Captured mirror ball]{
			\includegraphics[height=0.21\textwidth]{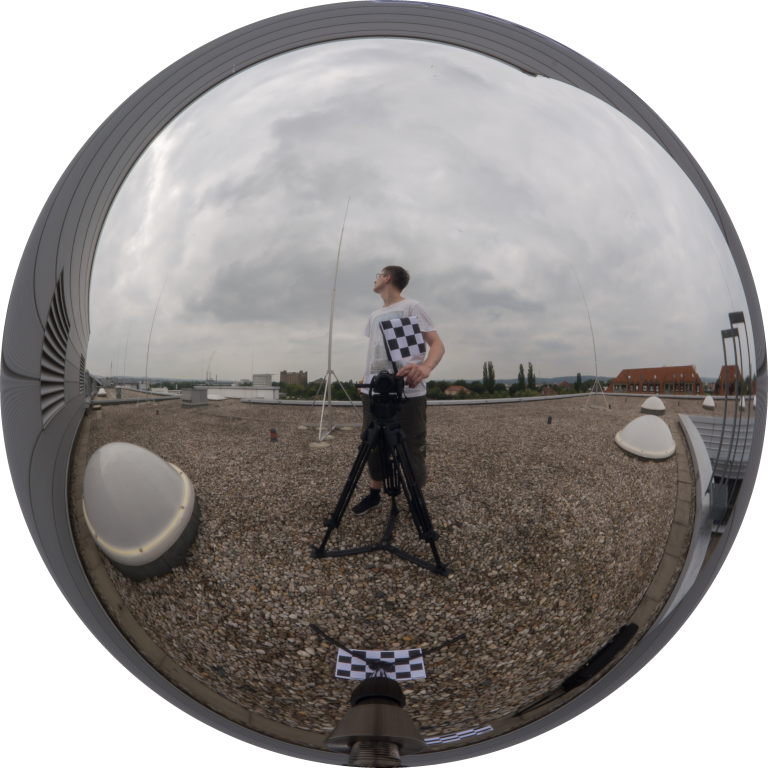}
			\label{fig:ball_result}
		}
		\hfill
		\subfloat[Pole-point, classical term]{
			\includegraphics[height=0.21\textwidth]{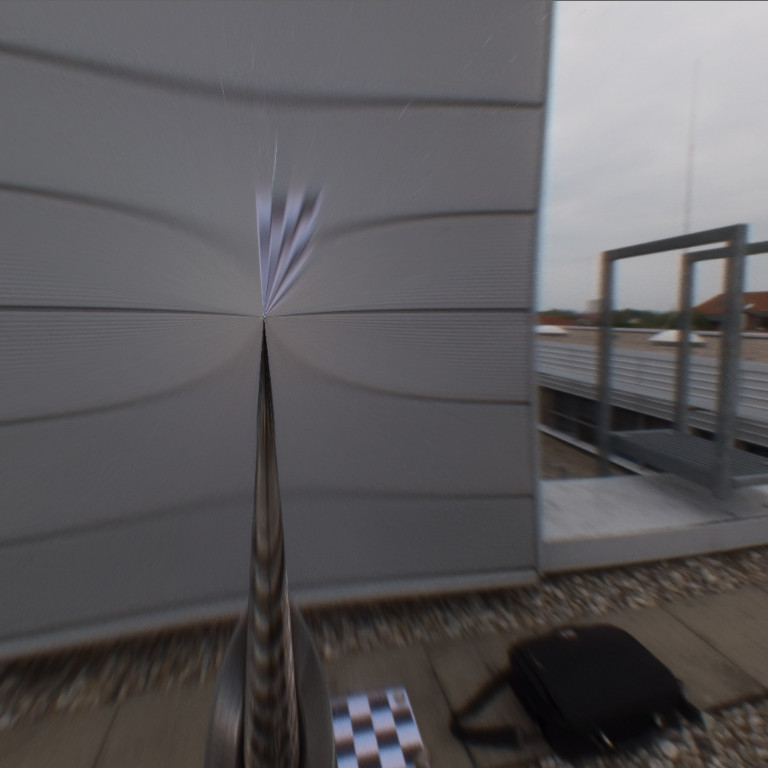}
			\label{fig:classic}
		}
		\hfill
		\subfloat[Pole-point, new term]{
			\includegraphics[height=0.21\textwidth]{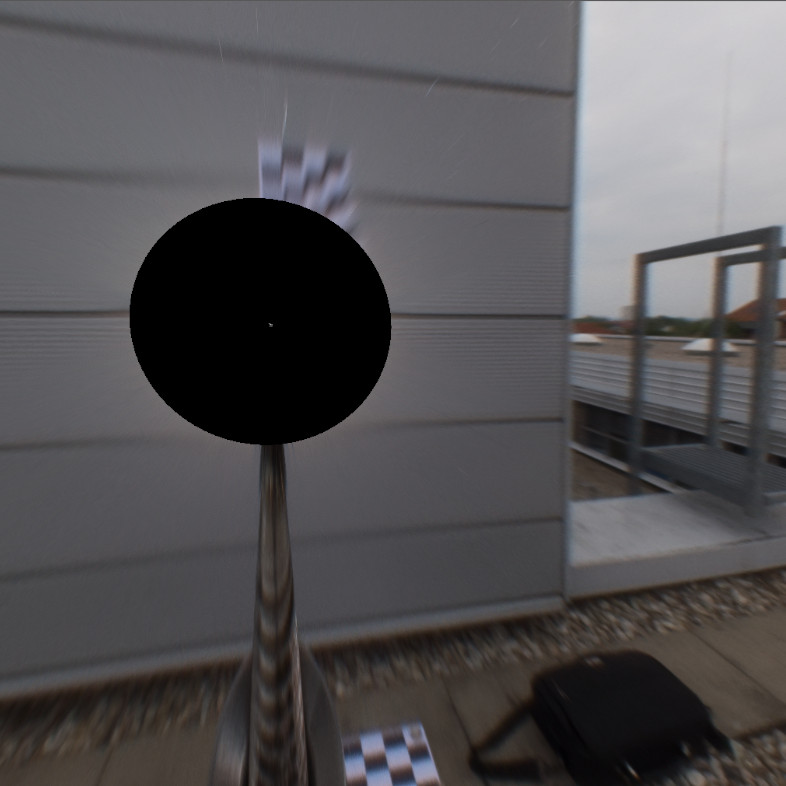}
			\label{fig:improved}
		}
	
		\caption{Setup to show the distortion correction}
		\label{fig:distort}
	\end{figure*}

	\section{The projection term}
	The mirror ball projection term, as used in this paper, maps the 3D reflection vector $\vec{r}$, as defined by the column vector $	\begin{bmatrix}
		r_x & r_y & r_z
	\end{bmatrix}^\intercal$, to the corresponding pixel on the 2D plane of an image, as defined by the column vector $\begin{bmatrix} {image}_x & {image}_y \end{bmatrix}^\intercal$. The origin of the coordinate system is at the center of the mirror ball, which itself is defined as a unit sphere, projecting onto the image plane as a unit disk.

	The term is a projection, which expresses a virtual camera-view at the center of an infinitely large sphere, allowing one to look around the full 360° environment, as reflected by the mirror surface of the imaged sphere.

	\subsection{Deriving the classical mirror ball projection term}

	The orthographic camera has its parallel view-rays defined by the incident ray column vector as written in Equation \ref{eq:cam}. As an orthographic camera captures the image, both the image's x,y-plane and the camera's x,y-plane are identical.
	\begin{equation}\label{eq:cam}
		\vec{i}=\begin{bmatrix}
			0 & 0 & 1
		\end{bmatrix}^\intercal
	\end{equation}
	The law of reflection for a light ray hitting a smooth surface is defined by Eq. \ref{eq:law}, with $\vec{r}$ being the reflected ray, $\vec{n}$ the surface normal and the dot '$\cdot$' being the scalar product.
	\begin{equation}\label{eq:law}
		\vec{r}=2\left(\vec{i}\cdot\vec{n}\right)\vec{n}-\vec{i}
	\end{equation}
	As per definition, the surface normal at each point of a unit sphere with its center at the origin, equals the vector as drawn from the origin to the surface at each point. With the orthographic camera's definition Eq. \ref{eq:cam}, the normal vector and the incident ray share the same xy-plane, leaving just the depth component of the normal $n_z$ as an unknown. Using that information, we can populate the law of reflection Eq. \ref{eq:law} and express the reflection vector $\vec{r}$ with the image's xy-plane:
	\begin{equation}\label{eq:insert}
		\left[\begin{matrix}r_x\\r_y\\r_z\\\end{matrix}\right]=2\left(\left[\begin{matrix}0\\0\\1\\\end{matrix}\right]\cdot\left[\begin{matrix}{image}_x\\{image}_y\\n_z\\\end{matrix}\right]\right)\left[\begin{matrix}{image}_x\\{image}_y\\n_z\\\end{matrix}\right]-\left[\begin{matrix}0\\0\\1\\\end{matrix}\right]
	\end{equation}
	Simplifying Eq. \ref{eq:insert}, it expresses this system of equations:
	\begin{equation}\label{eq:system}
		\left\{\begin{matrix}r_x=2n_z\ {image}_x\\r_y=2n_{z\ }{image}_y\\r_z=2{n_z}^2-1\ \ \ \ \ \\\end{matrix}\right.
	\end{equation}
	However, we need this system the other way around. We want to get the correct image pixel, based on the reflection direction $\vec{r}$. Solving the 3rd equation of Eq. \ref{eq:system} for $n_z$, gives us
	\begin{equation}\label{eq:substitute}
		n_z=\ \sqrt{\frac{r_z+1}{2}}
	\end{equation}
	Finally, we may substitute $n_z$ as defined by Eq. \ref{eq:substitute} into equation 1 and 2 of system Eq. \ref{eq:system} Simplifying the substitution gives us our mirror ball projection term:
	\begin{equation}\label{eq:simple_project}
		\begin{bmatrix} image_x \\ image_y \end{bmatrix}=\frac{1}{\sqrt{2(r_z+1)}}\begin{bmatrix} r_x \\ r_y \end{bmatrix}
	\end{equation}

	\subsection{Extending the projection term}
	The classical projection term Eq. \ref{eq:simple_project} suffers from distortion around the pole-point, as seen in Fig. \ref{fig:classic}, because it assumes an orthographic camera. A perspective camera obscures part of the mirror sphere as a function of sphere radius and distance from the center. The true pole-point of the projection is thus not actually captured. This issue may be side-stepped by use of another geometries, such as basing the mirror on a parabola, as shown in \cite{hyperbolic}. Staying with the mirror ball, the projection can also be adjusted to account for the mis-match between mathematical model and captured real-life mirror ball, by defining the blind spot inherit to capturing the mirror ball with a perspective camera.

	The missing information can be conceived as the cone-shaped 'solid angle', defined by $360^{\circ} -\alpha$, $180^{\circ}<\alpha<360^{\circ}$, where $\alpha$ expresses the field of view, that the mirror sphere is reflecting from the perspective of the camera's image. Any mapping from $\vec{r}$ going outside of the sphere's view cone will remain undefined. We need to remap the existing information to only fall within the visible cone. This can be achieved by stretching the reflection ray's image mapping by scalar $\sin{\left(\frac{\alpha}{4}\right)}$. This ratio stems from the amount of surface representing the reflected information decreasing by the sine of the reflected angle divided by 4, as we go from the middle point of the sphere to its edges. By stretching the reflection rays and allowing them to become undefined when going past the image plane's unit circle, we rigidly define the missing information. This shows up as a black circle in place of the classical mirror ball projection's pole-point, depicted in Fig. \ref{fig:improved}. Note how the lines of the wall are parallel, as they are in real-life with the correction of the improved term. Including said stretching scalar, results in the new and improved mirror ball projection term, shown in Equation \ref{eq:improved_project}. This term has been implemented in a demo WebApp, which can be accessed alongside sample photos and video footage on GitHub: \href{https://github.com/FrostKiwi/MirrorBall}{github.com/FrostKiwi/MirrorBall}
	\begin{equation}\label{eq:improved_project}
		\begin{bmatrix} image_x \\ image_y \end{bmatrix}=\frac{1}{\sqrt{2(r_z+1)}\sin{\left(\frac{\alpha}{4}\right)}}\begin{bmatrix} r_x \\ r_y \end{bmatrix}
	\end{equation}

	\begin{figure*}
		\subfloat[Setup, view from above]{
			\includegraphics[height=0.23\linewidth]{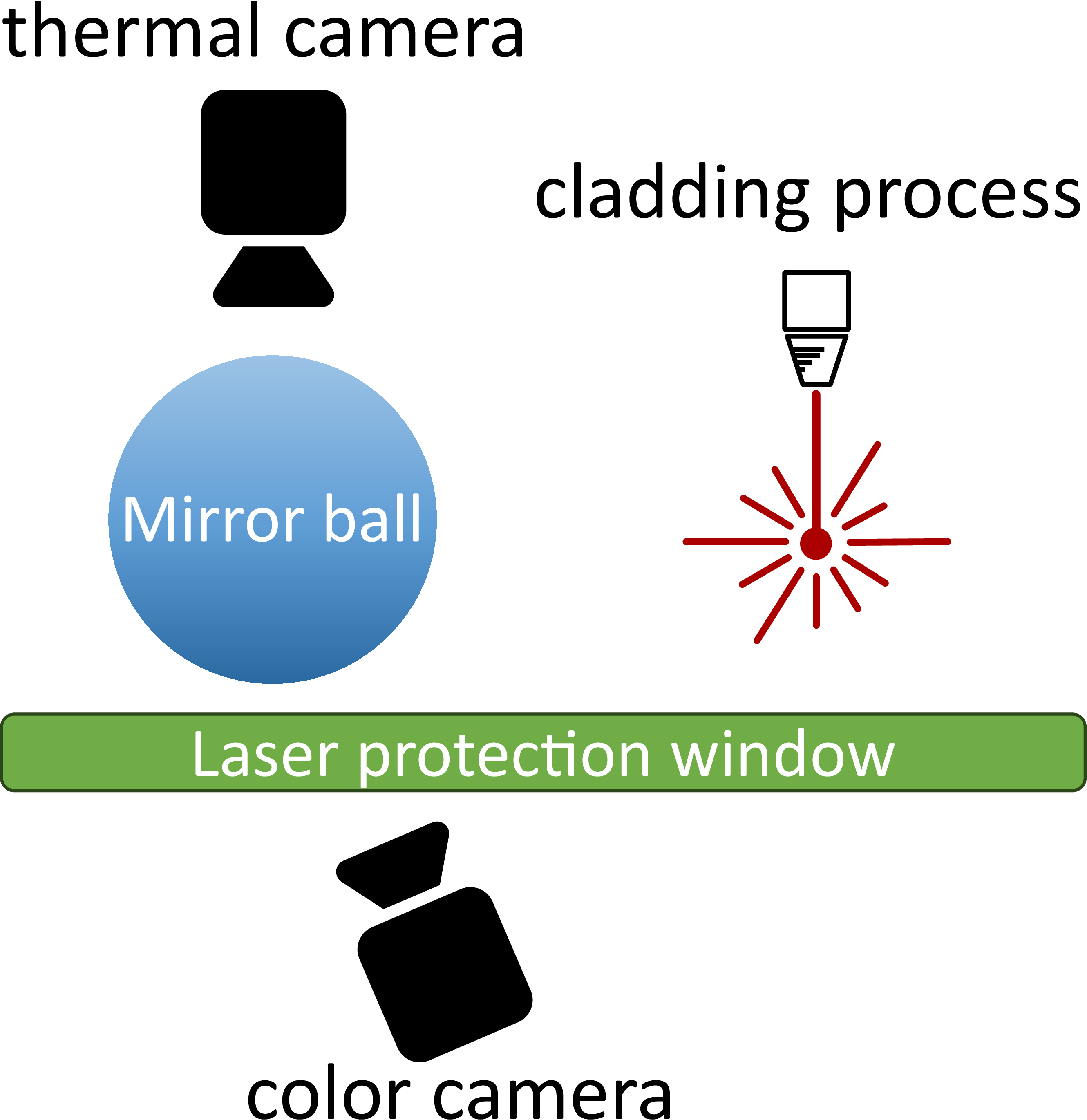}
			\label{fig:setup}
		}
		\hfill
		\subfloat[Outside view]{
			\includegraphics[height=0.23\linewidth]{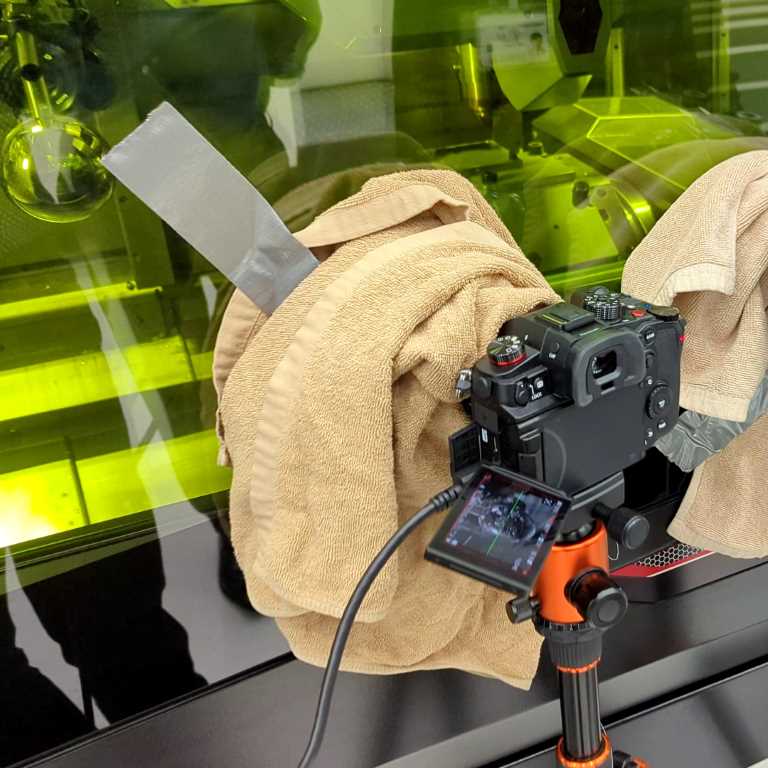}
			\label{fig:outside}
		}
		\hfill
		\subfloat[Chrome steel ball, \diameter 100mm]{
			\includegraphics[height=0.23\linewidth]{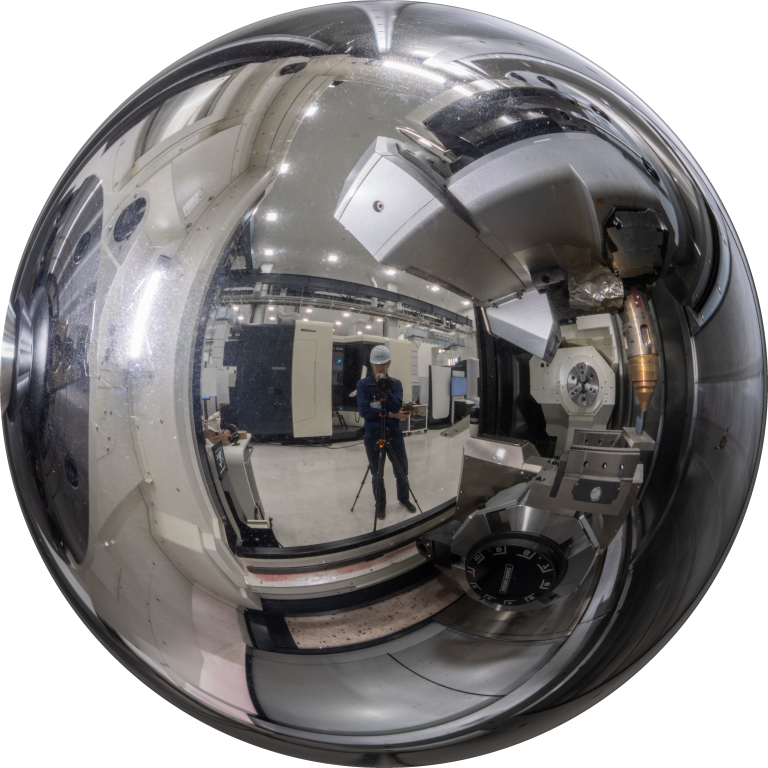}
			\label{fig:100mm}
		}
		\hfill
		\subfloat[Stainless steel ball, \diameter 50mm]{
			\includegraphics[height=0.23\linewidth]{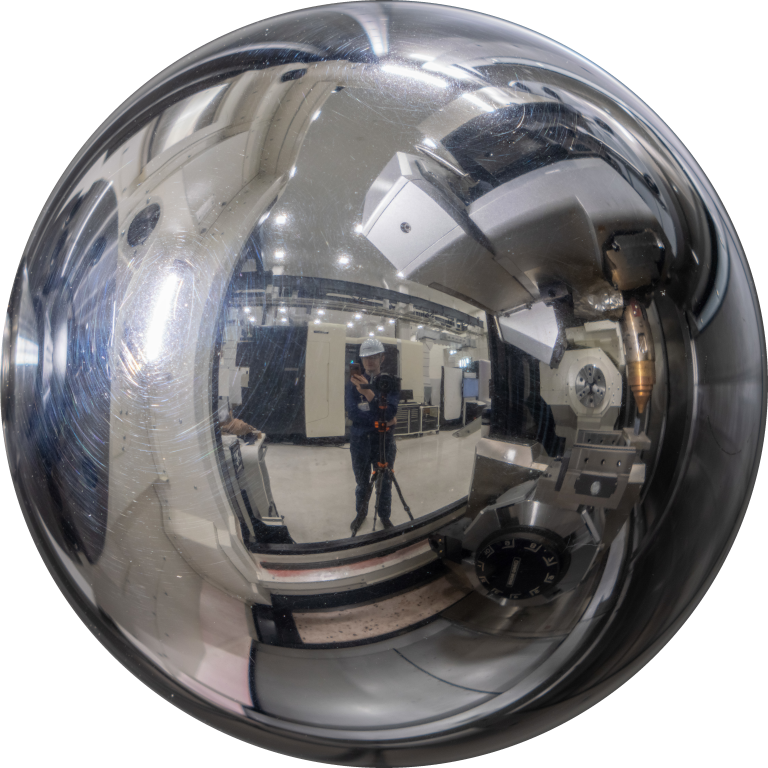}
			\label{fig:50mm}
		}
		\caption{Setup to determine viability of the mirror ball projection for process monitoring}
		\label{fig:monitoring}
	\end{figure*}

	\section{Process monitoring use-case}
	A laser cladding process inside a DMG MORI 'LASERTEC 3000 DED hybrid' industry machine was monitored via the mirror ball projection, to determine its efficacy for process monitoring and its ability to merge multiple camera sources into one coherent environment projection. A color camera was placed outside the machine, capturing mirror ball through the machine's protective glass. A thermal camera was placed inside the machine, opposite the color one, as visualized by the schematic Fig. \ref{fig:setup}. 
	
	Captured were a $3k^2$ pixel resolution color live-stream and combined with a low-resolution infrared stream to augment the monitoring experience. A further $4k^2$ px resolution color-only live-stream has been made to compare the impact of resolution. Snippets from both the $4k^2$ px and combined source live stream have been made available in the demo WebApp to be viewed by the reader and are also attached as supplementary material.

	\subsection{Quality evaluation}
	As expected from the mirror ball projection, the resolution towards the pole-point is weak, but everywhere else shows a well defined image, with enough resolution to properly judge the state of the machine. Even so, fine adjustments of the machine's tools cannot be reliably performed based on this view, as the resolution stretched across the projection's $360^{\circ}-\alpha$ field of view does not leave enough resolution to judge distances on the millimeter scale. Increasing the resolution does not necessarily constitute a solution, since the ball's surface blemishes also gain increased prominence. One the flip-side, bringing a bigger sphere too close to the monitoring target influences the result via parallax, as discussed in \ref{parallax}.
	
	Additionally two mirror balls of different size have been captured without the protective glass in the way, shown in Fig. \ref{fig:100mm} and \ref{fig:50mm}. Based on the model, the smaller the ball, the wider its field of view, when captured with a perspective camera. In reality, a ball which is too small needs a rather advanced lens to be properly focused. Futhermore, surface blemishes and scratches throw off the camera's auto-focus. The projection of the mirror ball \ref{fig:50mm} shows a softness to the projection, because the auto-focus accidentally focused on the scratches, instead of the environment in the  mirror image. On the flip-side, the projection of the larger ball \ref{fig:100mm} shows stronger parallax distortion, making image registration less reliable. Flipping between the two shows how the ball size changes the projection in slight ways, leading to an alignment which is not perfect.

	\subsection{Multi-source registration}
	Multiple camera views capturing the same mirror ball projection can be rotated into each other by a 3x3 rotation matrix, achieving image registration. Fig. \ref{fig:color} and \ref{fig:thermal} show the resulting mirror ball video feeds during laser cladding. Note, how both feeds show the hot metal sparks in different parts of the reflection. After correcting for the different rotations, both feeds are merged and can be switched between, as shown in figures \ref{fig:color_proj} and \ref{fig:thermal_proj}. Alignment settings become invalid as soon as a camera is moved though, leading to a fragile setup. Even though the camera positions are different, the projections are at least mathematically in the exact same spot, thus having no parallax, according to the model.
	
	\subsection{Parallax}\label{parallax}
	In reality the mirror balls have a physical size, thus their reflections originate from the surface, creating a new kind of parallax, easily observed by viewing the mounting point of the mirror ball, which shows strong distortion. Without Z-depth data, this cannot be corrected for. Whether the parallax is important enough to influence this use-case is entirely dependant on the balance between ball size and distance to the monitoring target. As a general rule, targets further away than the ball's diameter, preserve their shape in the projection well, though this a subjective statement. Finally, aligning based on one target leads to mis-alignment in the rest of the projection, an effect more pronounced with increasing ball size.

	\section{Conclusion}
	This paper extended the classical mirror sphere projection term by a new distortion scalar $\sin{\left(\frac{\alpha}{4}\right)}$ and its field of view parameter $\alpha$. This parameter allows one to express the distortion numerically, that is introduced by a perspective camera, when capturing a mirror ball projection. By counteracting this distortion, a more accurate registration of multiple camera outputs was achieved, when capturing the same mirror ball projection from different viewpoints.

	Monitoring a process via a mirror ball has highlighted the fragility of the setup and will not produce better quality results when compared to a wide-angle camera, installed at the spot of the mirror ball. Whether the drawbacks of the mirror ball setup can justify it's use, will depend on the difficulty of setting up a more traditional vision system. It is a viable option, but only becomes a favorable one in the niche use-case of a camera not being installable at the monitoring spot.

	\bibliography{references}
\end{multicols}
\end{document}